\documentclass{IOS-Book-Article}

\usepackage{mathptmx}
\usepackage{soul}\setuldepth{article}
\usepackage{amsmath}
\usepackage{amssymb}
\usepackage[colorinlistoftodos, textwidth=3cm]{todonotes} 
\usepackage{comment}

\def\hb{\hbox to 11.5 cm{}}

\newtheorem{definition}{Concept Explication.} 
 
\newtheorem{example}{Example}

\newcommand{\sforall}{\widetilde{\forall}}

\begin{document}

\pagestyle{plain}
\def\thepage{}
\begin{frontmatter}              

\title{Know your exceptions}

\subtitle{Towards an Ontology of Exceptions in Knowledge Representation}
\author[A, B]{\fnms{Gabriele} \snm{Sacco}\orcid{0000-0001-5613-5068}%
\thanks{Corresponding Author: Gabriele Sacco, Free University of Bozen-Bolzano, Piazza Domenicani 3, Bolzano, BZ, 39100, Italy; 
   E-mail: gsacco@fbk.eu.
   }},
\author[B]{\fnms{Loris} \snm{Bozzato}\orcid{0000-0003-1757-9859}}
and
\author[A]{\fnms{Oliver} \snm{Kutz}\orcid{0000-0003-1517-7354}}

\runningauthor{G. Sacco et al.}
\address[A]{Free University of Bozen-Bolzano, Italy}
\address[B]{Fondazione Bruno Kessler, Italy}

\begin{abstract}
Defeasible reasoning is a kind of reasoning where some generalisations may not be valid in all circumstances, that is general conclusions may fail in some cases. Various formalisms have been developed to model this kind of reasoning, which is characteristic of common-sense contexts. However, it is not easy for a modeller to choose among these systems the one that better fits its domain from an ontological point of view.\\
In this paper we first propose a framework based on the notions of exceptionality and defeasibility in order to be able to compare formalisms and reveal their ontological commitments. Then, we apply this framework to compare four systems, showing the differences that may occur from an ontological perspective.
\end{abstract}

\begin{keyword}
Exceptions\sep Ontology\sep 
Defeasible Reasoning\sep Non-Monotonic Logics\sep Knowledge Representation
\end{keyword}
\end{frontmatter}
\markboth{February 2024\hb}{February 2024\hb}

\section{Introduction} \label{intro}
Imagine being a modeller who is developing an ontology to represent knowledge about the zoological domain. Part of your knowledge are the propositions ``\textit{birds fly}'' and ``\textit{carnivora eat meat}''. But, you know also that Tweety is a bird who does not fly (in fact she is a penguin) and that Po is a carnivorous who does not eat meat (because he is a panda). The question now is: how can we model this knowledge in our ontology?
In fact, it seems reasonable wanting to not reject the information about birds and carnivora, since they are useful to provide knowledge regarding many individuals belonging to those classes. 
Therefore, we would like to model both the general properties of classes and the particular knowledge regarding the atypical individuals.

These kinds of situations have been studied in the context of knowledge representation and reasoning as cases of \emph{defeasible reasoning}.\footnote{In this paper we choose to use ``defeasible'' as the general and non-technical term to refer to that kind of reasoning where you can retract some conclusions. Other terms used in the literature, like ``default'', when used here, refer to the specific formalisms, like in the case of \textit{default logic}.} Defeasible reasoning is defined as a kind of reasoning which ``fails to `preserve truth' under all circumstances'' \cite[p. 94]{Bochman2018}.
Using the examples above, the conclusion that Po eats meat is retracted when we come to know that Po is a panda bear, which mostly eat plants.

The study of defeasible reasoning in the context of Artificial Intelligence (AI) led to the development of many logic-based accounts with the goal of modelling this kind of reasoning, called \emph{non-monotonic logics}. ``Non-monotonic'' means exactly that adding new knowledge may lead to retract some conclusions, or, in other words, that some conclusions may not hold in all possible circumstances. However, these accounts rely on different intuitions and assumptions and so they may behave differently when it comes to their application to a specific domain. More specifically, they `behave differently' for example in the way they spell out notions of `possible circumstance' and thus what specific conclusions might or might not be supported by the formal system. 
This means that, without making those intuitions and assumptions explicit, it is difficult to choose the system that suits best from an ontological point of view the specific domain the modeller is interested in capturing.

Therefore, the goal of our work is to develop a framework which can be used as a common ground for the comparison and consequently spelling out the ontological commitment of the systems for modelling defeasible reasoning. In this way, we have some criteria 
to guide the choice of formalisms when it comes to decide which one fits better for a specific domain modelling.  
To show this, we individuate the ontological commitments of the four reference defeasible systems discussed in \cite{BRACHMANchap11}, through a systematic comparison and ontological discussion based on such framework.  The first account examined is \emph{closed world assumption} \cite{Reiter77CWA, Reiter80CWA+DC}, which is considered to be the first, most straightforward, formalised way of dealing with defeasible reasoning. The next two systems can be seen as different directions for refining closed world assumption. In the first of these two refinements, the method develops the idea of minimising the extension of predicates, leading to the account developed by John McCarthy, namely \emph{Circumscription} \cite{MCCARTHY198027}. 
In the second refinement of the closed world idea, an approach is developed based on the idea of adding only information which is consistent with the knowledge base, leading to the system developed originally by Raymond Reiter, namely  \emph{Default logic} \cite{Reiter80DefaultLogic}. 
Finally, we include an approach developed by Robert C. Moore, namely \emph{Autoepistemic logic} \cite{Moore85AutoepistemicLogic}, which tries to bring together the advantages of circumscription and default logic. 
We start our study from the analysis of these key
approaches since they define the foundation of 
more recent non-monotonic approaches in KR (for example,
the non-monotonic extensions of description logics~\cite{GIORDANO15FromPLtoDL, BonattiLW06_CircumscriptionInDLs, SaccoBK23FormalApproach}), thus  providing 
a solid ground for a wider, more comprehensive study.

In the classical, pre-Carnapian view of logic, it was often assumed that formal logic does not make any ontological commitments \cite{Carnap37LogicalSyntaxOfLanguage,hyper2010}. And even when, following Carnap, we admit different logics, speaking of ontological commitments may still be considered implausible if we assume that each logic is agnostic with respect to the specific domain it is used to model and thus the choice of logic is merely considered a question of expressivity and formalisation style \cite{DBLP:conf/jowo/ToyoshimaN21,DBLP:conf/fois/BorgoPT14}.
Against this, we here hold that  through the choices made in developing a non-monotonic formalism, a theory about defeasibility and exceptionality is conveyed and endorsed. Therefore, in order to make the ontological commitments explicit, and to understand the consequences of such choices, we need firstly a framework which allows us to extract and formulate such a foundational theory.

The central notions in our proposed framework are  \emph{defeasibility} and \emph{exceptionality}. In particular, our analysis of the four systems for defeasibility will be guided by an articulation of the ways in which these two properties are represented and interact. Thus, we start by developing a definition of ``exception'' in Section~\ref{what_is_an_exception}. In Section~\ref{defeasibility_and_exceptions}, we discuss the relation between exceptions and defeasibility through the introduction of the key distinction between \emph{universal generalisations} and \emph{defeasible generalisations}. In Section~\ref{formal_models_of_defeasible_reasoning}, we introduce briefly the four defeasible systems in order to provide the basis for the comparison in Section~\ref{comparison}. We conclude with Section~\ref{conclusions} by summarising the results and discussing some future directions for the research.

\section{What Is an Exception?} 
\label{what_is_an_exception}

In this section, we discuss what it means to be an exception and propose a definition of this notion. We start with a description of sources of defeasibility for knowledge\footnote{Note that here the concept of ``knowledge" is used in a non-strict and non-technical way. In fact, from a philosophical point of view, talking of ``defeasible knowledge" is contradictory since knowledge cannot be false by the very definition of the term, e.g.\ in the tradition explicating knowledge as variants of ``justified true belief". 
    Therefore, we should speak of justified belief rather than full-fledged knowledge: however, for the sake of simplicity, we prefer to use the term ``knowledge" even if we ask to keep in mind this important observation.}, that is, a distinction between different types of knowledge that are used to explain why an entity may be considered exceptional. 

    \subsection{Sources of Defeasibility} \label{sources_of_defeasibility}

        The first reason for having exceptions to something we know is that the knowledge we have is \emph{incomplete}: 
        we do not have enough knowledge to explain away all the exceptions. 
        In this case, we have a problem of \emph{completeness}, we lack elements that could make the general picture coherent.\\
        Another feature of our knowledge that makes it defeasible is \emph{uncertainty}. In situations where knowledge is uncertain, we accept that there may be circumstances in which what we know does not really apply. 
        In this case, the problem is \emph{justification}, that is, the fact that we have reasons justifying our knowledge which support it but do not affirm it completely. 
        Different from uncertain knowledge is \emph{vague} knowledge. The difference is that in the case of uncertain knowledge we are able to clearly distinguish those circumstances in which our knowledge correctly applies and those in which it does not. In the presence of vagueness, instead, we are not able to fix precisely the circumstances in which what we know applies. 
        Here the issue is that of \emph{meaning}: the borders of a concept are not clearly fixed. 
        The last case we consider of knowledge admitting exceptions is that of \emph{simplified} knowledge. Differently from incomplete knowledge, we may know that precisely speaking there are no exceptions, but for the sake of simplicity we accept some as such. In other words, we may be able to make our knowledge precise enough that it encompasses also those cases which at first glance seem exceptions, however it would make the knowledge too complex and so we prefer to have a less precise but simpler version of it.  
        In this case the problem is the \emph{precision}, we accept less precision for the sake of simplicity.
        
        These four types of knowledge are not mutually exclusive, for example we can have knowledge which is both uncertain and incomplete. However, these distinctions will help us in identifying those contexts in which we need to pay particular attention to the possibility of defeasible knowledge, and thus to the presence of exceptions. 
        
    \subsection{Common-Sense Definitions of ``Exception''} \label{common-sense_definitions}
    In order to begin a delimitation of the concept of exception, 
    we collected four common-sense definitions 
    for the term ``Exception'' 
    that we can find in dictionaries.
    {\footnotesize
        \begin{description}
        \item[Cambridge dictionary:\footnotemark]
        ``someone or something that is not included in a rule, group, or list or that does not behave in the expected way'';\footnotetext{\url{https://dictionary.cambridge.org/dictionary/english/exception}}

        \item [Oxford languages:\footnotemark]
        ``a person or thing that is excluded from a general statement or does not follow a rule'';
\footnotetext{\url{https://www.oed.com/dictionary/exception_n}}
        
        \item [Collins dictionary:\footnotemark] 
        ``An exception is a particular thing, person, or situation that is not included in a general statement, judgment, or rule.''.
        \footnotetext{\url{https://www.collinsdictionary.com/dictionary/english/exception}}
    \end{description}
    }
    \noindent
    All the above definitions rely on two main concepts for defining what an exception is: some kind of generalisation, expressed through terms like ``rule'', ``group'', ``list'', ``general statements''; and an instance or individual which does not fit to or is excluded from that generalisation. Even if we endorse the approach of setting up our definition in these terms, we clearly need to ontologically refine the underlying ideas.
    
    \subsection{Exceptions, Counter-Examples and Errors} \label{exceptions_counterexamples_errors}
    To carry out our refinement of this core idea, we propose to analyse how the relation between the generalisation and the individual can take place. First of all, what we need to stress is the fact that the exclusion of the individual happens in spite of the fact that we expect it to be included. For example, if we are speaking of flying birds, a dog cannot be considered an exception, since we do not expect it to satisfy that generalisation.\\ 
    However, there is still the possibility that we are not speaking of exceptions even in that case. In fact, assume that we have a generalisation and an instance to which that generalisation should apply, but it actually does not, that is, it is an instance which is excluded from that generalisation. This discrepancy can happen in three cases:
    {\footnotesize
        \begin{description}
            \item[Counter-example:] the generalisation is falsified by the instance, e.g.:\\ 
            \textit{The school can be reached only by 
            bus or train}, but \textit{Bob} arrives by car.
            
            \item[Error:] the instance is faulty, e.g.:\\ 
            \textit{Students have a unique ID card number},
            but \textit{Alice} is registered with a wrong, non-unique number.
            
            \item[Exception:] both the generalisation and the instance are correct, e.g.:\\ 
            \textit{Students have no salary},
            but \textit{Eve} is a PhD student with a bursary.
        \end{description}
        }
    As we can see, according to the truth-values we give to the propositions describing the generalisation and the instance we have three different situations. The peculiarity in the case of exceptions is that even if the two propositions contradict each other, we want to consider both of them true. The reason for this can be recognised in the kind of knowledge we are using, namely whether it is incomplete, uncertain, vague or simplified knowledge. However, this means that we cannot rely on classical deductive reasoning, since this would lead to a contradiction.

    \smallskip\noindent
    We now collected the basic pieces that we can use to elaborate a definition of exceptions. Firstly, we need to be in a particular context where the knowledge we have is incomplete, uncertain, vague or simplified. Secondly, we need two elements for individuating exceptions, a generalisation and an individual that for some reason is excluded from that generalisation, even if it should be included. Finally, we want to consider both the knowledge about the generalisation and the particular individual as correct. Therefore, we give our explication of the notion of `exception' as follows: 
    
    \begin{quotation} \label{Working definition}
        An \textbf{exception} is an individual justifiably excluded from a generalisation, without causing a contradiction. 
        
%
    \end{quotation}
%
%
This explication is our starting point for the discussion in the next sections, which will lead to a more refined version of this position.
    
\section{Defeasibility and Exceptions} \label{defeasibility_and_exceptions}
To begin refining the definition of exceptions we can start from one of the main elements that compose it, that is the notion of generalisation. We begin the analysis by looking at some relevant fields where generalisations that admit exceptions have been studied. 

    \subsection{Defeasible Generalisations, Generics and \emph{Ceteris Paribus} Laws} \label{generics_ceteris-paribus_laws_defaults}
    We can start from the distinction made in \cite{BRACHMANchap11} between \emph{universals} and \emph{generics}, even if slightly re-elaborated: if in \cite{BRACHMANchap11} those terms are used for properties, we will instead use them for propositions
    in order to be coherent with the working definition developed above in Section~\ref{Working definition}. In particular, universal propositions correspond to generalisations such as \textit{Dogs are animals}, that is, where the predicated property applies to all the instances of the first class, whereas generic propositions correspond to those like our examples above, where there could be instances that do not instantiate the predicated property, such as Po the panda bear for \emph{Carnivora eat meat}.
    
    \subsubsection{Generics}
    An interesting insight regarding this distinction comes from the literature on \emph{generics}~\cite{sep-generics, DefaultGenerics}. Generics are defined as sentences reporting a regularity regarding particular facts that can be generalised even if they admit exceptions. In the penguin example, for instance, the sentence \textit{Birds fly} can be considered a generic since it states something about the individuals of the class \textit{bird}, but still allows exceptions like the members of the sub-classes \textit{penguins}, \textit{ostriches} and \textit{emus}. The other peculiarity beyond tolerating exceptions is that in this kind of generalisations we do not have information about how many individuals satisfy the generalisation.\\
    Since we are interested in generalisations that admit exceptions in general, including those indicating quantities like \textit{most books are paperback} or \textit{no car is able to fly}, we do not restrict ourselves to considering just generic generalisations. For this reason, we will refer to those generalisations more broadly as \emph{defeasible generalisations}, highlighting their capacity to admit exceptions and thus  fail to apply to all the cases. However, research of linguists, philosophers and cognitive scientists in generics suggests that the distinction between \emph{universal generalisations} and defeasible ones may not be only a methodological distinction but that they are really distinct phenomena, at least in the sub-kind of generics \cite{Leslie07Generics}.
    
    \subsubsection{\emph{Ceteris paribus} laws}
    Another related topic is that of \emph{ceteris paribus} laws (henceforth \textit{cp-laws}), which attains the general fields of philosophy of science and metaphysics. Considering scientific fields such as biology, psychology or economics, it has been noted that most of the laws they state are not universal. That is, in particular sciences ``there are (actual and merely possible) situations in which the above generalisations do not hold, although all the conditions obtain that are explicitly stated in the antecedents of these generalizations'' \cite{sep-ceteris-paribus}. These laws have been termed \emph{ceteris paribus}, which is a Latin expression that can be translated as ``all other things being equal'': 
    as we can read from the above quotation,
    such laws are strictly related to the notions we are considering in our
    analysis on defeasibility in terms of generalisations and exceptions.
    The idea behind cp-laws is that they are valid assuming that other factors that may influence their application to a specific case remain constant or are completely absent. Determining the precise meaning of this is 
    the goal of studies on cp-laws:
    for the time being, we are more interested in some insights we can gain for our purposes from such studies.
    
    First of all, we note that
    the generalisations as we have considered them until now can be seen as weak laws we apply everyday in common-sense reasoning: in this sense, cp-laws as laws of the particular sciences are in the mid-way between common-sense generalisations and the generalisations made, for instance, in fundamental physics, which are considered paradigmatic universal laws. For this reason, the insights coming from the study of cp-laws with respect of their peculiar capability of admitting exceptions can \emph{a fortiori} be applied to defeasible generalisations which are weaker generalisations not refined by the process of scientific inquiry.
    A first interesting observation concerns exclusive cp-laws, that is, those for which interfering factors have to be \emph{absent}, which can be distinguished between \emph{definite} and \emph{indefinite} cp-laws \cite[Section 3.2]{sep-ceteris-paribus}. The former are those for which the interfering factors excluded from the antecedent are specified, that is, it is possible to complete the cp-law with those factors. Intuitively, indefinite cp-laws are those for which it is not possible to do such a completion if not in the form of adding to the antecedent the clause ``excluding all the interfering factors'' without specifying what these interfering factors are.\\
    We return to this point in the next section because the possibility of completing the cp-law suggests that it may be possible to do the same for defeasible generalisations and this allows to transform a defeasible generalisation into an equivalent universal generalisation through the addition of this complement.

    The last observation comes exactly from the possibility of the transformation suggested above. In fact, exclusive cp-laws are subject to a dilemma \cite[Section 4]{sep-ceteris-paribus}: if we transform a cp-law into a universal law, it will tend to be false, since it is implausible that all the interfering factors have been considered and even only one exception would turn into a counter-example to a universal law. Note that a similar observation is raised in \cite{BRACHMANchap11} with respect to the strategy of listing all the exceptions to a generalisation.\\ 
    On the other side, if we add a completer like that for indefinite exclusive cp-laws that is similar to ``assuming that nothing interferes'', the risk is to end up with a trivially true statement given its unwelcome similarity to a statement like ``All As are Bs or not all As are Bs". This is particularly interesting from the perspective of a modeller, since even if we avoid transforming a defeasible generalisation into a tautology, it would remain very moderately informative about the actual conditions under which the generalisation should apply and under which it should admit exceptions instead.

    In the next section, we will discuss the relationship between universal and defeasible generalisations in light of the features discussed above and especially of the dilemma emerging for cp-laws.
    
    \subsection{Universal and defeasible generalisations} \label{universals_and_generics}
    First of all, we can fix some terminology. In a classical formal setting, like \emph{first order logic} (FOL), we can represent only the first kind of generalisation, the universal one. This is done through a universal quantifier and a conditional:
    \begin{definition}
        A \textbf{universal generalisation} is formally represented through the FOL formula $\forall x (Px \rightarrow Qx)$, with the standard interpretation of ``all the Ps are Qs''.
    \end{definition}
    Therefore, to refer to universal generalisations we can also use the term \emph{$\forall$-generalisation}.\\ 
    Since what characterises the generalisation is the quantifier, 
    we can represent defeasible generalisations by substituting the quantifier with something different.  
    Therefore, we denote defeasible generalisations 
    with the symbol $\sforall$
    and refer to them with the term \emph{$\widetilde{\forall}$-generalisations}.\footnote{Note that a similar move is done to represent syntactically generics: the quantifier \emph{Gen} is introduced for generic sentences \cite{sep-generics}. Because we do not want to commit ourselves with generics specifically, we will not use \textit{Gen}, but instead use the symbol $\widetilde{\forall}$.}
    Consequently, the aim of non-monotonic logics can be seen as introducing a formal way to model $\widetilde{\forall}$-generalisations. As we have said above, there are many ways this has been done, and we will analyse them below. A first \emph{naive} intuition that can be adopted is \emph{listing} all the exceptional cases and excluding them \cite{BRACHMANchap11}. However, since most $\widetilde{\forall}$-generalisations are like indefinite cp-laws for which it is not possible to be sure that all the exceptions have been considered, this is not a good strategy and, as we mentioned above, we are exposed to the first horn of the dilemma \ref{dilemma}.\\ 
    So, as \cite{BRACHMANchap11} points out, we end up saying something like \emph{Carnivora eat meat except those that do not}, and the real goal becomes characterising somehow ``those that do not''. In other words, the goal becomes to characterise the class of \emph{exceptions} without falling on the second horn of the dilemma, since the impression is that now the defeasible generalisation is trivial. Nevertheless, this problem gives us a first insight into a relation between  $\forall$-generalisations and $\widetilde{\forall}$-generalisations.\\ 
    In fact, the idea here is that a defeasible generalisation can be seen as a universal generalisation where some instances are removed from the generalisation, namely the exceptions to that generalisation considered as universal. 
    Consequently, we can exploit this intuition to precisify the meaning of $\widetilde{\forall}$-generalisations semi-formally: 
    \begin{definition}
        A \textbf{defeasible generalisation} is formally represented as $\widetilde{\forall} x (Px \rightarrow Qx)$ and it is interpreted as ``a strict subset of the $P$s is also a subset of the $Q$s''.
    \end{definition}
    The meaning of ``strict subset'' is defined by each particular formal system.\\ 
    The relation between universal and defeasible generalisations can be summarised as
        ``$\widetilde{\forall}$-generalisation $\Longleftrightarrow$ $\forall$-generalisation $-$ exceptions''
    where $\Longleftrightarrow$ means an equivalence between the two formulations. The idea here is that we can see a defeasible generalisation as a universal generalisation where we remove from the range of universal quantifier the exceptional individuals.\\
    For instance, consider the case of the statement: \textit{Prime numbers are odd numbers}. There is a clear exception which is the prime number 2. We can apply the equivalence above and write
        ``$\widetilde{\forall}x (x\ \text{is a prime number} \rightarrow x\ \text{is an odd number}) \Longleftrightarrow \forall x (x\ \text{is a prime number} \rightarrow x\ \text{is an odd number}) - \{2\}$''.
    This was a simple case where we know for sure that there is only one exception, however, we can see that in more complex cases the key point of $\widetilde{\forall}$-generalisations represented as universal ones is how to characterise the class of exceptions. In fact, according to the definitions of these two kinds of generalisations we can see that the complement of the subset identified by the quantifier $\widetilde{\forall}$ is the set of the exceptions. In this sense, the second horn of dilemma \ref{dilemma} should be made precise by saying that a formulation like the second term in the equivalence is trivial only if we are not able to individuate the exceptions to that generalisation.\\
    Therefore, the framework we proposed suggests that the crucial point of a model for defeasible reasoning is to develop a method able of individuating the correct, or better, the reasonable exceptions to be given as a complement of the defeasible generalisation to make it universal.\\ 
    Related to this, another observation is relevant: the class of exceptions is relative to the generalisation we are considering and we cannot possibly have exceptions without a generalisation. Consequently, in principle, it cannot be defined \textit{a priori}, that is before the statement of the generalisation. For instance, in the case of Tweety, you cannot exclude from the class of birds the exceptional ones, without first saying that we are considering the defeasible generalisation \textit{Birds fly}. \\
    From this observation, it emerges also that for each class, we can have more than one class of exceptions, exactly because we can have an instance which is exceptional with respect of a generalisation but not another. In the case of birds, for example, there are birds which are exceptional because they do not fly, like the penguins or the ostriches, and others which are exceptional because they do not use nests, like the cuckoos. In the first case, penguins and ostriches are exceptional birds because they are exceptions to the generalisation \textit{Birds fly}, in the second case, cuckoos are exceptional birds because they are exceptions to the generalisation \textit{Birds build nests}.
    \subsection{Characterising Exceptions} \label{characterising_exceptions}
    In the previous sections we set the ground for comparing the different non-monotonic logics in order to have insights on the commitments they make which could be relevant for a modeller who would like to use them. The key point emerged in the latter subsection: given the relationship between universal and defeasible generalisations and since the aim of non-monotonic logics is to represent defeasible generalisations, we can compare them according to how they try to \emph{characterise the exceptions}.\\
    According to this framework we can propose a new understanding of exception:
    \begin{definition} \label{framework_definition}
        An \textbf{exception} is an individual belonging to a \emph{justified} subset of the domain which if explicitly excluded from the scope of the quantifier of a false $\forall$-generalisation transforms it in a true $\sforall$-generalisation.
    \end{definition}
    
    To confirm whether this is a good definition we have to check if it allows to recognise all the relevant features we attributed to exceptions so far.\\
    Firstly, individuating exceptionality as a relation between an individual and a generalisation is obviously preserved. The fact that it is excluded from that generalisation is also preserved since the idea is exactly that the subset to which the exception belongs is considered as the set of completers, which if mentioned in the generalisation allows to individuate the exact subset in the scope of the defeasible generalisation. It also captures the idea that exceptions do not falsify the generalisation, since they do not falsify the defeasible generalisation, but they still have the property contradicting the generalisation since they are excluded from the subset instantiating the generalised property.\\
    However, the definition above leaves unspecified the idea of justification for the exclusion of the exceptions and of the different kinds of knowledge as their sources: these are exactly those factors that allow to individuate actual exceptions, which is the goal of the specific formalisations. Therefore, we can see this definition as incomplete: studying how each non-monotonic logic specifies this idea of justification is exactly our interest.\\
    In the next section, we will explain four formalisms considering their key aspects in order to see how they can complete the definition of exception above, and thus to understand what it means in each case to be an exception, and therefore how defeasibility is understood. 

\section{Formal Models of Defeasible Reasoning} \label{formal_models_of_defeasible_reasoning}
Here we give a sketch of the approaches in order to make the comparison in the next section understandable. We provide a brief introduction to the approaches in Appendix~\ref{appendix}
and for a more precise and formal discussion we refer to \cite{BRACHMANchap11}. 

\subsection{Closed World Assumption} \label{closed-worlds_assumption}
    The central idea of the CWA is that what is not explicitly mentioned can reasonably be assumed to be false. So this leads to defeasibility because you could come to know something new, which changes it from being considered false to being considered true. However, CWA is a defeasible system in a broad sense. In fact, it does not represent defeasible knowledge at the object level, rather the modeller can update manually the KB and so change information in it. So to say, with the CWA we are always in the default state of knowledge and we cannot really represent non-default cases.\\
    For instance, if you have a knowledge base containing the sentences ``Tweety flies'', ``Tweety is a bird'' and ``Chilly is a bird''. Since you do not have the sentence ``Chilly flies'', you assume as valid its negation, that is ``Chilly does not fly''. If then you add to your knowledge base ``Chilly flies'', you remove that assumption. 

\subsection{Circumscription} \label{circumscription}
    The key idea of circumscription is introducing a class of \emph{abnormality} predicates $Ab_i$, which is minimised. Therefore, we can represent $\widetilde{\forall}$-generalisation as 
    $\forall x ((P(x) \land \lnot Ab(x)) \rightarrow Qx)$
    which can be interpreted as ``if $x$ is a $P$ and is not abnormal, then it is a $Q$ too".\\
    So, in this case we have two kinds of generalisation even if they are not distinguished at the object level. That is, it is not introduced a new form of generalisation, but only a new predicate, which indirectly makes the generalisations where it is present to behave differently. In terms of Definition \ref{framework_definition} above, the justification here is given by being recognised as abnormal. \\
    In fact, are considered exceptions those individuals which necessarily are abnormal. This is due to the mechanism of reducing the extension of $Ab$ as much as possible, which represent the assumption that we are always in the most normal possible condition. In fact, in circumscription we order the interpretations according to the extension of the $Ab$ predicates, where the smaller the better.\\
    For example, consider the knowledge base containing the propositions:
        $\forall x (Bird(x) \land \lnot Ab(x) \rightarrow Flies(x))$, $Bird(tweety)$, $Bird(chilly)$, $\lnot Flies(chilly)$, $(tweety \neq chilly)$. 
    Chilly is correctly considered abnormal, since it does not fly and we can conclude that Tweety flies because we consider the minimal models, and a model where $Ab$ contains only Chilly is more normal than one containing also Tweety.

 \subsection{Default Logic} \label{default_logic}
    The intuition behind default logic is to add a set of rules that allow to specify which assumptions should be made, given that they are consistent. A rule has the following form: $\langle \alpha:\beta/\delta \rangle$ where $\alpha$ is a sentence corresponding to the \emph{prerequisite} for the application of the rule, $\beta$ is one that corresponds to the \emph{justification} for the application of the rule and $\gamma$ is the \emph{conclusion} of the rule. The idea is that if we know $\alpha$ and $\beta$ is consistent with what we know, we can conclude $\delta$.\\
    These rules are the ones that allow to represent defeasible generalisations. In fact, default rules do not necessarily apply to all individuals, but only to those which are consistent to do so. Therefore, exceptions to a default rule are those individuals for which the justification does not apply. Interestingly, this is an inverse perspective with respect to our view above, since in this case the justification is needed for applying the defeasible generalisation, rather than to individuate the exceptions.\\
    For instance, consider the knowledge base containing the propositions: $Bird(tweety)$, $Bird(chilly)$, $\lnot Flies(chilly)$ and the rule $Birds(x):Flies(x)/Flies(x)$.
    In this case we have only a rule which says that if something is a bird and it is not inconsistent to assume that it flies, then we can conclude that it flies. We can see that this is true only for Tweety, since we know that Chilly does not fly and so we have in our expansion that Tweety flies.
    
\subsection{Autoepistemic Logic} \label{autoepistemic_logic}
    In autoepistemic logic, the intuition is that of trying to exploit introspection, represented with the introduction of an unary connective which is used to represent what is believed. This allows to represent defeasibility because adding or removing beliefs affect the beliefs I believe, in this way the conclusions that I reach by exploiting what I know to believe can change. That is, we can exploit the indexicality of the introspection operator \cite{sep-logic-nonmonotonic}.\\
    In fact, defeasible generalisations are modelled through the use of the introspection operator in the antecedent of a conditional. For this model we use a ``higher level'' kind of beliefs which are those with the introspection operator. They are ``higher level'' in the sense that they recursively or indexically speak of other beliefs in the theory.\\
    We introduce the unary operator $\textbf{B}$ with the meaning of ``it is believed that...''.
    For instance, imagine our knowledge base containing the following sentences
           $Bird(tweety)$, $Bird(chilly)$, $(tweety \neq chilly)$, $\lnot Flies(chilly)$, 
           $\forall x (Bird(x) \land \lnot \textbf{B} \lnot Flies(x) \rightarrow Flies(x))$. 
        From this KB, since we have $\lnot Flies(chilly)$ and  $\lnot \textbf{B} \lnot Flies(chilly)$, we have a $ \textbf{B} \lnot Flies(chilly)$  because there is no way of entailing $\lnot Flies(x)$. Therefore, we cannot infer $Flies (chilly)$ but we can infer $Flies (tweety)$.
    
\section{The Comparison} \label{comparison}

Now that we have an idea of the key points of these four theories we can start to compare them and show in which regard they differ. The comparison of the four formalisms is developed along four lines: if the defeasibility is represented through a \emph{syntactic} or a \emph{semantic} mechanism; if defeasibility is understood at the \emph{epistemic} or at the \emph{ontological} level; if exceptions are represented \emph{explicitly} or \emph{implicitly}; and finally, if the defeasible generalisations are understood as formulae representing propositions (at the \emph{logical} level) or as meta-language rules (at the \emph{meta-logical} level). 
In Table \ref{comparison-table}, we show an overview of the results of this comparison, which we will discuss in the next subsections.
\begin{table}
    \centering
    \begin{tabular}{|c|c|c|c|c|}
    \hline
         & \begin{minipage}{50pt}\centering
             \textbf{Semantic Vs Syntactic}
         \end{minipage}  & 
         \begin{minipage}{50pt}\centering
             \textbf{Epistemic Vs Ontological}
         \end{minipage} & 
         \begin{minipage}{50pt}\centering
             \textbf{Explicit Vs Implicit}
         \end{minipage} & 
         \begin{minipage}{50pt}\centering
             \textbf{Logical vs Meta-Logical}
         \end{minipage}\\ 
    \hline
    CWA & Syntactic & Epistemic & / &  / \\
    Circumscription & Semantic & Ontological & Explicit & Logical \\
    Default logic & Syntactic & Epistemic & Implicit & Meta-Logical \\
    Autoepistemic logic & Syntactic & Epistemic & Implicit & Logical \\\hline
    \end{tabular}
    \caption{Summary of the comparison between the four approaches.}
    \label{comparison-table}
\end{table}

    \subsection{Syntactic or Semantic Approach} \label{syntaxVSsemantics}
    The first and more evident distinction between the approaches is that between a syntactic or semantic representation of the defeasibility of these approaches. That is, if $\widetilde{\forall}$-generalisations have been modelled through a semantic or syntactic mechanism. This is relevant, because if defeasibility is modelled through a syntactic mechanism, then we do not need to intervene on the ontology. On the other hand, it means also that we do not have any ontological notion of exceptions. The opposite can be observed for semantically modelled defeasibility.\\
    As we have said above, having CWA in this comparison is not completely fair, since we do not really have a representation of defeasibility in the system. However, it may be interesting for some discussion that we can have regarding this system. In fact, even if the CWA is a semantic assumption, because the assumption concerns the world, the mechanism used is syntactical, since what counts is the presence or not of literals in the KB.\\
    Regarding circumscription, the situation is more simple since we introduce a new predicate with an interpretation that uses an order on models. This makes it a semantic approach for defeasibility. Default and autoepistemic logic, instead, are both syntactical systems. In fact, the default rules do not have an interpretation but they work on the basis of the notion of syntactical consistency. Whereas, the introspection operator has not a semantic interpretation and so the entire non-monotonic mechanism relies on a syntactic basis.
    
    \subsection{Epistemic or Ontological Level} \label{epistemicVSontological}
    The second important distinction concerns the source of the defeasibility, that is if the system represents defeasibility as something due to how the world is or to how we represent it. In the former case, we are considering defeasibility as something occurring at the ontological level, while in the latter we are considering it as something at the level of knowledge, that is epistemic. 
    This distinction seems related to the previous one, in fact, in this comparison they overlap. However, since the distinctions are in principle different, since the former concerns the type of mechanism used for modelling defeasibility, whereas the latter considers at which level defeasibility emerges and is understood, we maintain them separately and will investigate their correlation as future work. 
    In CWA, despite its name, defeasibility comes from our way of knowing, since the assumption is that we know all the relevant knowledge. Therefore, here we are at the epistemic level. 
    In circumscription, we are at the ontological level. In fact, defeasibility comes from the fact that we recognise abnormal individuals, to which generalisations regarding normal circumstances do not apply. 
    In default logic, the interpretation of the KB, composed of the set of FOL sentences and the set of default rules, is that of a theory and the key notion is that of consistency among the beliefs entailed by the theory. Therefore, here we are treating defeasibility as something occurring at the epistemic theory. 
    In the case of autoepistemic logic, it is not immediate the level at which we are considering the defeasibility. In fact, at the object language we are working with beliefs, but the new introspection operator is not a predicate, therefore it is not at the object level. So, since the defeasible mechanism is realised through the definition of extensions like for default logic, the predominant level is that of epistemology.
    
    \subsection{Explicit or Implicit Representation of Exceptions} \label{explicitVSimplicit}
    The third line of comparison which allows to distinguish the approaches regards how exceptions are represented in the system, more precisely if they are represented explicitly or implicitly. 
    Having in the ontology a class (or classes) of exceptions is a relevant issue.    
    In fact, on one hand, having explicit exceptions in the model would mean that exceptional entities would be immediately recognisable and can be treated also with respect of their being exceptions. On the other hand, this would also mean that the ontology should be structured in order to accommodate the property of being an exception.\\ 
    In this distinction and the next one, it emerges the peculiarity of CWA. In fact, since there are no defeasible generalisations, in CWA there is not a notion of exception and \textit{a fortiori} there is not a notion of implicit of explicit exception.\\ 
    Circumscription, instead, is probably the easiest case for this distinction: in fact, in this formalism exceptions are those individuals which necessarily are abnormal. This is due to the mechanism of reducing the extension of $Ab$ as much as possible. Therefore, in this system exceptions are represented explicitly as those individuals which are in one of the abnormal predicates.\\
    In default logic, exceptions are identified by inconsistency. That is, the instances which are exceptions are those which satisfy the prerequisite, but not the justification and therefore, assuming them would lead to an inconsistency. However, these instances are not marked in some way, so to recognise them we have to consider the default rules and look for those individuals which satisfy the prerequisite but not the justification. Consequently, exceptions here are represented implicitly.\\
    The case of autoepistemic logic is similar to that of default logic: we can consider exceptions those beliefs about individuals which do not correspond to the introspection status expressed through the operator in the antecedent. This means that they satisfy the classical part of the antecedent but not the one with the introspective operator, which can be seen respectively as the analogous of the prerequisite and the justification of a default rule. Therefore, autoepistemic logic represents exceptions implicitly too.
    
    \subsection{Logical or Meta-logical}
    The last comparison regards the form given to defeasible generalisations, that is if they are modelled at the logical level and so we can reason on them as we do with classical FOL formulas or if they are outside the language and so they are meta-logical. This distinction is interesting because it allows us to distinguish the systems where it is possible to exploit the representations of the defeasible generalisations to reason on them and those where it is not possible to reason about them.\\
    Again, this distinction cannot be applied to CWA since it does not represent defeasible generalisations. For what regards circumscription instead, defeasible generalisations are modelled at the logical level, whereas in default logic, as suggested above, they are considered rules at the meta-logical level.\\
    More interesting is the case of autoepistemic logic. In this case, defeasible generalisations are formulae in the language, therefore they are treated at the logical level. However, they do not have a standard extensional semantics since the introspection operator gains its meaning through the notion of stable expansion, which is a syntactic notion. Therefore, even if autoepistemic logic can be classified as modelling defeasible generalisations as formulae about which we can reason, it does so in a peculiar way with respect of circumscription.
    
\section{Conclusions and Future Works} \label{conclusions}
In this paper, we introduced a framework based on the notions of generalisation and exception for a comparison of non-monotonic logics that should help a modeller in choosing the better approach according to her needs for a specific domain. In Section~\ref{what_is_an_exception} we analysed and discussed what it means to be exceptional from a broader point of view. This discussion supplied the basis for a more precise discussion in Section~\ref{defeasibility_and_exceptions} thanks also to the insights coming from the study of generics and ct-laws. In this section we developed the definitions for the key notions of defeasible generalisation and exception for the framework. Then, after having described the formal approaches used in the comparison in Section~\ref{formal_models_of_defeasible_reasoning}, we did the comparison in Section~\ref{comparison}, which is summarised in Table~\ref{comparison-table}.\\
As future works, we plan to develop further the framework for example through a comparison with \cite{schlechta16NewPerspective}, where an idea of \emph{generalised quantifier} is discussed for representing defeasible reasoning.
Moreover, we want to extend the comparison to other kinds of non-monotonic logics like \emph{preferential} approaches \cite{KLM90, LEHMANN1992} and \emph{modal} approaches \cite{DefaultGenerics}, but also to applications of non-monotonic logics in the context of \emph{description logics} like in \cite{GIORDANO15FromPLtoDL, BritzEtAlKLMinDLs}. Another interesting discussion we would like to develop is the possibility of representing defeasible reasoning through monotonic logics. Finally, we plan to exploit the analysis developed here regarding the notions of exception and generalisation to refine our formal system outlined in \cite{SaccoBK23FormalApproach}.
\bibliography{biblio}
\bibliographystyle{vancouver}

\newpage
\appendix
\section{Formal Models of Defeasible Reasoning: extended} \label{appendix}
    \subsection{Closed World Assumption}
    The central idea of the CWA is that what is not explicitly mentioned can reasonably be assumed to be false. So this leads to defeasibility because you could come to know something new, which change it from being considered false to being considered true. However, CWA is a defeasible system in a broad sense. In fact, it does not represent defeasible knowledge at the object level, rather the modeller can update manually the KB and so change information in it. So to say, with the CWA we are always in the default state of knowledge and we cannot really represent non-default cases.\\
    To obtain this result, the basic strategy is to define a new consequence relation $\models_C$\footnote{For the notation we use as reference \cite{BRACHMANchap11}.} based on an expanded knowledge base (KB$^{+}$) where the negation of all the atomic formulae which are not mentioned in the original knowledge base (KB) are added:
    \begin{definition}
        $KB \models_C \alpha$ if and only if $KB^+ \models \alpha$, where $KB^+ = KB \cup \{\lnot p\ |$ p is atomic and $KB \nvDash p\}$
   \end{definition}    
    This definition works in the propositional case, however for the complete application of the intuition behind CWA in FOL we need to extend it with what is called \emph{domain closure}:
    \begin{definition}
        $KB \models_CD \alpha$ if and only if $KB^{\diamond} \models \alpha$, where $KB^{\diamond} = KB^+ \cup \{\forall x[x=c_1 \lor ... \lor x=c_n]\}$ with $c_1, ... , c_n$ all the constant symbols present in the KB.
    \end{definition}
    The idea is that we restrict our domain to only those individual which are named in the KB.\\
    Now, if we move this approach in our framework, we can see that, first of all, this approach is formulated in propositional logic first. This means that we do not really have a notion of generalisation. But even in the FOL case, where we add the domain closure, we do not have two forms of generalisations. Therefore, we do not really have a notion of exception in this models. But this means that defeasibility in this case is not a represented, but, rather, it can be considered as a meta-property of the approach. In fact, if we want to modify our knowledge, thus making it defeasible, we have to materially change the knowledge base.\\
    Consequently, this system does not really allows to represent defeasible knowledge, consider the example below:
    \begin{example}
        Consider the knowledge base $KB = \{B(t),\ \forall x (B(x) \rightarrow F(x)\}$, which represent the knowledge that Tweety is a bird and that birds fly.\\ 
        $KB \models \{F(t), B(t)\}$, therefore $KB^{+} = KB = \{B(t),\ \forall x (B(x) \rightarrow F(x)\}$ and so $KB^{\diamond} = \{B(t),\ \forall x (B(x) \rightarrow F(x), \forall x (x = t)\}$.\\
        However, if we want to model the fact that Tweety is not able to fly and therefore the defeasibility of ``Birds fly'', namely $\forall x (B(x) \rightarrow F(x)$, we cannot simply add $\lnot F(t)$ to the $KB$, since it would lead to an inconsistency, as shown below.\\    
        $KB = \{B(t),\ \forall x (B(x) \rightarrow F(x),\ \lnot F(t)\}$.\\
        $KB \models \{F, \lnot F, B\}$\\
        Since all the atoms are entailed by $KB$, $KB^{+}$ is equal to $KB$ again:
        $KB^{+} = KB = \{B(t),\ \forall x (B(x) \rightarrow F(x),\ \lnot F(t)\}$ and so $KB^{\diamond} = \{B(t),\ \forall x (B(x) \rightarrow F(x), \forall x (x = t), \lnot F(t)\}$\\
        In fact, as said above, in CWA we cannot really represent defeasible knowledge.
        The same apply in the case of FOL with domain closure.        
    \end{example}
    Therefore, CWA is not a system for defeasible reasoning \textit{strictu sensu} and so also in our comparison it should be consider not at the same level to the others.
    
    \subsection{Circumscription} 
    The key idea of circumscription is introducing a class \emph{abnormality} predicates $Ab_i$, which is minimised. Therefore, we can represent $\widetilde{\forall}$-generalisation as 
    \[\forall x ((P(x) \land \lnot Ab(x)) \rightarrow Qx)\]
    which can be interpreted as if something is a P and is not abnormal then it is a Q too.\\
    So, in this case we have two kinds of generalisation even if they are not distinguished at the object level. That is, it is not introduced a new form of generalisation, but only a new predicate, which indirectly makes the generalisations where it is present to behave differently. In terms of definition \ref{framework_definition} above, the justification here is given by being recognised as abnormal. \\
    In fact, are considered exceptions those individuals which necessarily are abnormal. This is due to the mechanism of reducing the extension of $Ab$ as much as possible, which represent the assumption that we are always in the most normal possible condition. In fact, in circumscription we order the interpretations according to the extension of the $Ab$ predicates, where the smaller the better. 
    \begin{definition}
        $\mathcal{I}_1 \leq \mathcal{I}_2$ if and only if for every $Ab_i$ predicate $\mathcal{I}_1(Ab) \subseteq \mathcal{I}_2 (Ab)$
    \end{definition}
    Then, we define a new consequence relation imposing that something is entailed if and only if it is entail by the minimal models satisfying the $KB$: 
    \begin{definition}
        $KB \models_{\leq} \alpha$ iff for every interpretation $\mathcal{I}$ such that $\mathcal{I} \models KB$, either $\mathcal{I} \models \alpha$ or there is a an $\mathcal{I}'$ such that $\mathcal{I}' < \mathcal{I}$ and $\mathcal{I}' \models KB$.
    \end{definition}
    Consequently, defeasibility is understood as presence of abnormal entities for which a generalisation does not apply, that is there are generalisations which regards only normal entities.\\
    In order to make things clearer, consider the example below described in \cite{BRACHMANchap11}
    \begin{example}
        Consider the knowledge base 
        \begin{align*}
           KB = \{\forall x (Bird(x) \land \lnot Ab(x) \rightarrow Flies(x)),\ Bird(tweety),\ Bird(chilly),\\ \lnot Flies(chilly),\ (tweety \neq chilly)\}. 
        \end{align*}
         Chilly is correctly considered abnormal, since it does not fly and we can conclude that Tweety flies because we consider the minimal models, and a model where $Ab$ contains only Chilly is more normal than one containing also Tweety.
    \end{example}
    
    \subsection{Default Logic} 
    The intuition behind default logic is to add a set of rules that allow to specify which assumptions should be made, given that they are consistent. A rule has the following form: $\langle \alpha:\beta/\delta \rangle$ where $\alpha$ is a sentence corresponding to the \emph{prerequisite} for the application of the rule, $\beta$ is one that correspond to the \emph{justification} for the application of the rule and $\gamma$ is the \emph{conclusion} of the rule. The idea is that if we know $\alpha$ and $\beta$ is consistent with what we know, we can conclude $\delta$.\\
    These rules are the ones that allow to represent defeasible generalisations. In fact, default rules do not necessarily apply to all the individuals, but only to those which is consistent to do so. Therefore, exceptions to a default rule are those individual for which the justification does not apply. Interestingly, this is an inverse perspective with respect to our view above, since in this case the justification is needed for applying the defeasible generalisation, rather than to individuate the exceptions.\\
    To make the intuition working formally, it is defined the notion of \emph{extension}. Given a default theory $(\mathcal{F}, \mathcal{D})$ where $\mathcal{F}$ is the set of first order sentences and $\mathcal{D}$ is the set of default rules, we can define an extension.

    \begin{definition}
        A set of sentences $\epsilon$ is an \textbf{extension} of $(\mathcal{F}, \mathcal{D})$ if and only if for every sentence $\pi$, $\pi \in \epsilon$ iff $\mathcal{F} \cup \{ \delta | \langle \alpha:\beta/\delta \rangle \in \mathcal{D}, \alpha \in \epsilon, \lnot \beta \notin \epsilon\} \models \pi$
    \end{definition}

    Therefore, sentence is in the extension of a default theory if and only if it is entailed by the first order sentences of the theory together with a suitable set of default rules. A suitable default rule is one that has the prerequisite in the expansion but not the negation of the justification.\\
    To understand better how default logic works, consider the example below from \cite{BRACHMANchap11}:

    \begin{example}
        Consider the default theory $\mathcal{F} = \{Bird(tweety),\ Bird(chilly),\ \lnot Flies(chilly) \}$ and $\mathcal{D} = \langle Birds(x):Flies(x)/Flies(x) \rangle $\\
        In this case we have only a rule which says that if something is a bird and it is not inconsistent to assume that it flies, then we can conclude that it flies. We can see that this is true only for Tweety, since we know that Chilly does not fly and so we have in our expansion that Tweety flies.
    \end{example}
    
    \subsection{Autoepistemic Logic}
    In autoepistemic logic, the intuition is that of trying to exploit introspection, represented with the introduction of an unary connective which is used to represent what is believed. This allows to represent defeasibility because adding or removing beliefs affect the beliefs I believe, in this way the conclusions that I reach exploiting what I know to believe can change. That is, we can exploit the indexicality of the introspection operator \cite{sep-logic-nonmonotonic}.\\
    In fact, defeasible generalisations are modelled through the use of the introspection operator in the antecedent of a conditional. For this model we use a ``higher level'' kind of beliefs which are those with the introspection operator. They are ``higher level'' in the sense that they recursively or indexically speak of other beliefs in the theory.\\
    We introduce the unary operator $\textbf{B}$ with the meaning of ``it is believed that...''. The meaning of this operator formally is given through two elements. The first one is that of \emph{stable} set:
    \begin{definition}
        A set $\epsilon$ is a \textbf{stable} set if and only if 
        \begin{enumerate}
            \item if $\epsilon \models \alpha$, then $\alpha \in \epsilon$ 
            \item if $\alpha \in \epsilon$, then $\textbf{B}\alpha \in \epsilon$
            \item if $\alpha \notin \epsilon$, then $\lnot \textbf{B}\alpha \in \epsilon$ 
        \end{enumerate}
    \end{definition}
    Formally, autoepistemic logic works in a similar way to default logic, in fact the second important notion is that of expansion. So, we can define a stable expansion
    \begin{definition}
        A stable set $\epsilon$ is a \textbf{stable expansion} if and only if for every sentence $\pi,\ \pi \in \epsilon$ iff $KB \cup \{\textbf{B}\alpha | \alpha \in \epsilon\} \cup \{\lnot \textbf{B}\alpha | \alpha \notin \epsilon\} \models \pi$
    \end{definition}
    To show how this works we can rely on an example.

    \begin{example}
        Our KB contains the following sentences 
        \begin{align*}
           Bird(tweety),\ Bird(chilly),\ (tweety \neq chilly),\ \lnot Flies(chilly),\\ 
           \forall x (Bird(x) \land \lnot \textbf{B} \lnot Flies(x) \rightarrow Flies(x)). 
        \end{align*}
        From this KB we have a stable expansion with $ \textbf{B} \lnot Flies(chilly)$ because we have $\lnot Flies(chilly)$ and  $\lnot \textbf{B} \lnot Flies(chilly)$ because there is no way of entailing $\lnot Flies(x)$. Therefore, we cannot infer $Flies (chilly)$ but we can infer $Flies (tweety)$.
    \end{example}





 
 

\end{document}